\documentclass[10pt,twocolumn,letterpaper]{article}

\usepackage{cvpr}              

\usepackage{booktabs}
\usepackage{multirow}
\usepackage[table]{xcolor}
\usepackage{graphicx}
\usepackage{xcolor}
\usepackage{bm}
\usepackage{amsmath}
\usepackage[normalem]{ulem}
\usepackage{float}
\usepackage{dsfont}
\usepackage{algorithm}
\usepackage{algpseudocode}
\usepackage{booktabs}
\usepackage{multirow}
\usepackage{array}
\usepackage{tabularx}
\usepackage{xcolor}
\usepackage{colortbl}

\definecolor{FDAMHeader}{HTML}{E8F1F8}
\definecolor{FDAMBest}{HTML}{E6DAEF}
\definecolor{FDAMSecond}{HTML}{FBE5D3}

\newcolumntype{C}{>{\centering\arraybackslash}X}

\DeclareRobustCommand{\bestval}[1]{%
    \begingroup
    \setlength{\fboxsep}{0.6pt}%
    \colorbox{FDAMBest}{\textbf{#1}}%
    \endgroup
}
\DeclareRobustCommand{\secondval}[1]{%
    \begingroup
    \setlength{\fboxsep}{0.6pt}%
    \colorbox{FDAMSecond}{\underline{#1}}%
    \endgroup
}

\makeatletter

\renewcommand{\footnoterule}{%
    \kern -3pt
    \hrule \@width 0.45\columnwidth
           \@height 0.4pt
    \kern 3pt
}

\newcommand{\cvprauthornote}{%
    \begingroup
    \renewcommand{\thefootnote}{}%
    \renewcommand{\@makefntext}[1]{%
        \noindent ##1%
    }%
    \endgroup
}

\makeatother

\definecolor{cvprblue}{rgb}{0.21,0.49,0.74}
\usepackage[pagebackref,breaklinks,colorlinks,allcolors=cvprblue]{hyperref}

\def\confName{CVPR}
\def\confYear{2027}

\title{FDAM: A Frequency-Direction-Aware Multi-Task Framework for \\ Pavement Crack Grid Localization}

\author{%
Haiyang Xu\textsuperscript{1} \quad Bin Pan\textsuperscript{1,*} \quad Mengze Wang\textsuperscript{1} \quad Qiaoying Qu\textsuperscript{1} \quad Zhenwei Shi\textsuperscript{2}\\[6pt]
{\small\textsuperscript{1}School of Statistics and Data Science, LEBPS, KLMDASR, LPMC, AAIS, NITFID, Nankai University}\\[2pt]
{\small\textsuperscript{2}Department of Aerospace Intelligent Science and Technology, School of Astronautics, Beihang University}
}

\begin{document}
\maketitle

\begin{abstract}
Pavement crack grid localization remains challenging for thin and tiny cracks due to weak visual cues and background interference.
Box detection often misses these cracks, while supervised semantic
segmentation faces limited pixel-wise annotations and high labeling
costs. In this paper, we propose a Frequency-Direction-Aware Multi-Task framework (FDAM) with box detection and grid classification branches. The core grid classification branch comprises Frequency-Domain Enhancement (FDE) and Directional Feature Refinement (DFR).
FDE adaptively weights spectral bands and fuses reconstructed features with spatial features. DFR captures thin crack patterns through multi-scale strip convolutions, direction-sensitive modulation, and feature recalibration. Sparse positive cells and ambiguous local predictions motivate adaptive grid supervision. Adaptive Importance Reweighting emphasizes uncertain predictions and local label transitions, with focal supervision for all cells. Our contribution is to couple the box detection and grid classification branches through shared feature learning for pavement crack grid localization. We validate FDAM on our RoadDistress2026 dataset of thin and tiny cracks and further evaluate it on the public datasets Crack500 and DeepCrack. The code is available at \url{https://github.com/Lab-PANbin/FDAM}.
\end{abstract}    
\section{Introduction}
\label{sec:introduction}

Pavement crack detection identifies surface cracks and their spatial
distribution for road condition assessment
~\cite{ayman2023recent,pan2023spatiotemporal}. We focus on pavement crack grid localization for thin and tiny cracks. We use thin to describe narrow apparent widths, whereas tiny
describes limited spatial extent relative to the image. These image-space characteristics, together with pavement textures and shadows, make reliable localization challenging
~\cite{fan2023pavement}.

Existing methods represent cracks using bounding boxes or pixel-wise masks. Transformer-based detectors predict categories and instance extents through set prediction
\cite{zhao2024detrs,huang2025real,huang2025deim,peng2025d}.
Convolutional segmentation aggregates multiscale features
\cite{zou2019deepcrack,yang2019feature,al2023asymmetric},
while lightweight and attention-based designs refine crack details
\cite{zhu2024lightweight,liang2024cnn,zhang2023integrated}.
Transformers \cite{liu2021crackformer,xu2022pavement,lin2023transcrack}
and state space models
\cite{gu2024mamba,liu2025scsegamba,gu2026crackssm,zhao2026mixercseg}
extend contextual modeling.
Multiscale and hybrid architectures integrate local detail with
broader context \cite{gu2022multi,guo2023pavement,xiao2023pavement,xu2026drcnet},
with further refinement through structural modeling and feature mixing
\cite{wang2024dual,han2024enhancing,zuo2024topology,zhang2024robust}.
Spectral processing provides complementary features for restoration
and segmentation
\cite{chu2023rethinking,zhang2024crack,yang2025ffr,joo2026frequency}.
Directional models encode oriented cracks \cite{chen2023devil},
while strip convolutions capture elongated objects in remote sensing
\cite{yuan2026strip}.

However, weak crack cues, costly annotations, and imbalanced supervision
still challenge thin and tiny crack localization.
Bounding boxes can contain substantial background, while fully
supervised segmentation requires precise pixel-wise masks.
Weakly supervised methods reduce annotation demands
\cite{huang2023weakly,konig2022weakly,inoue2023weakly,he2024research}
but still need to distinguish weak cracks from the background.
Although tailored losses and focal loss address foreground-background
imbalance
\cite{liang2024lightweight,siriborvornratanakul2023pixel,lin2017focal},
sparse positive grid cells and ambiguous label transitions remain
challenging. Reliable grid localization therefore requires stronger crack
representations and effective supervision across the image.

\begin{figure*}[t]
    \centering
    \includegraphics[width=\textwidth]{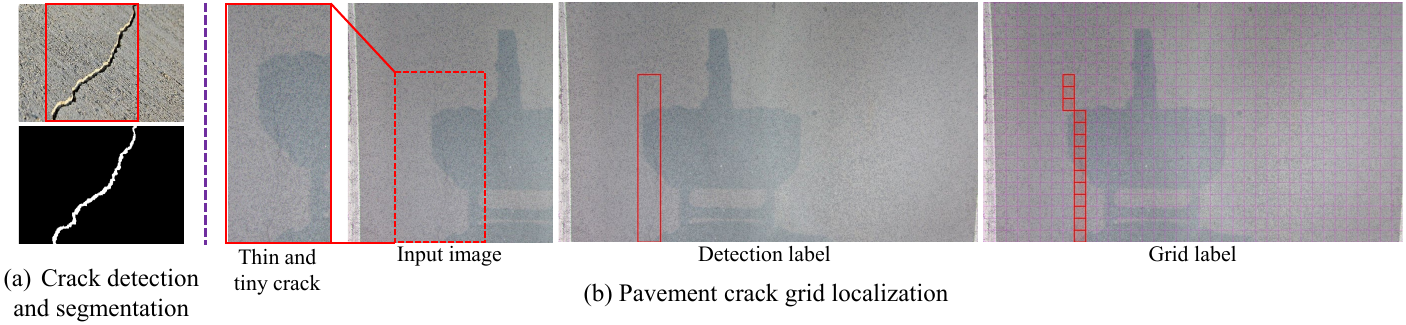}
    \caption{Illustration of pavement crack localization tasks and labels.
    (a) Crack detection and segmentation. Detection and pixel-wise
    segmentation labels are shown for a wide crack.
    (b) Pavement crack grid localization for thin and tiny cracks.
    From left to right: input image and crack detail, detection label,
    and grid label. The full-image grid defines regular classification
    units, and cells with highlighted boundaries indicate crack presence.}
    \label{fig:crack_grid_motivation}
\end{figure*}

Grid occupancy provides this local representation and has an engineering
basis in pavement assessment. China's \emph{Highway Performance Assessment Standard} (JTG~5210--2018) classifies longitudinal and transverse asphalt pavement cracks as low-severity when the main crack width does not exceed
$3~\mathrm{mm}$. Its automated assessment of the Pavement Surface Condition Index uses
occupied $0.1~\mathrm{m}\times0.1~\mathrm{m}$ grids to estimate distress
area~\cite{mot2018jtg5210}.
Inspired by this practice and prior joint box-grid learning~\cite{li2023grid}, we combine instance localization and local
occupancy prediction for thin and tiny cracks. Our grids operate in image coordinates without a prescribed physical size.
Fig.~\ref{fig:crack_grid_motivation} illustrates the task and its
complementary labels. 

To address this challenge, we propose FDAM, a
Frequency--Direction-Aware Multi-Task framework that couples box
detection and grid classification through shared feature learning.
A shared backbone and hybrid encoder support a transformer-based
detection branch~\cite{huang2025real} and the core grid classification branch. Within the grid branch, Frequency-Domain Enhancement (FDE) adaptively
weights spectral bands, while Directional Feature Refinement (DFR)
captures elongated crack structures. Adaptive Importance Reweighting emphasizes uncertain predictions and local grid-label transitions while retaining focal supervision over all cells. During inference, class-aware box-grid association filters grid predictions using detection categories and spatial overlap. Experiments on our newly constructed RoadDistress2026 dataset of thin and tiny cracks and the public Crack500 and DeepCrack datasets evaluate the framework and its components.

Our main contributions are summarized as follows:
\begin{itemize}
    \item We develop a multi-task framework for pavement crack grid
    localization of thin and tiny cracks through shared box and grid
    feature learning.

    \item We design a grid classification branch combining adaptive
    spectral weighting and directional feature refinement to strengthen
    weak crack representations.

    \item We introduce adaptive importance reweighting for grid focal
    supervision, emphasizing uncertain predictions and local label
    transitions while retaining supervision over all cells.
\end{itemize}
\section{Related Work}
\label{sec:related_work}

\subsection{Crack Detection and Segmentation}

Automated pavement analysis spans distress evaluation
~\cite{dong2021automatic}, crack localization~\cite{du2023modeling},
quantification~\cite{pan2023one}, and learned distress recognition
~\cite{dong2021deep}.
Transformer-based detectors predict object categories and boxes
through set prediction
~\cite{huang2025real,huang2025deim,peng2025d,zhao2024detrs}.
For pixel-wise crack prediction, convolutional networks aggregate
hierarchical and multiscale features
~\cite{al2023asymmetric,yang2019feature,zou2019deepcrack}.
Lightweight designs and attention-based integration improve efficiency
and feature representation
~\cite{li2024rethinking,liang2024cnn,zhang2023integrated,zhu2024lightweight}.
Transformers capture contextual dependencies through
attention~\cite{lin2023transcrack,liu2021crackformer,xu2022pavement}.
Selective state space modeling~\cite{gu2024mamba} also supports
long-range interactions in recent crack segmentation
networks~\cite{gu2026crackssm,liu2025scsegamba,zhao2026mixercseg}.

Preserving spatial detail remains essential for weak crack representations.
High-resolution multiscale features support semantic
segmentation~\cite{gu2022multi}, while crack-specific architectures
integrate local detail with broader context
~\cite{guo2023pavement,xiao2023pavement,xu2026drcnet}.
Other methods explore convolution-transformer integration~\cite{wang2024dual}, convolutional and state space
features~\cite{han2024enhancing}, topology-aware
modeling~\cite{zuo2024topology}, and global-local feature
mixing~\cite{zhang2024robust}.
These approaches advance dense crack prediction, but fully supervised
segmentation requires precise pixel-wise annotations.

\subsection{Frequency and Directional Feature Modeling}

Spectral and directional representations provide complementary cues
for recovering weak structures.
Frequency-domain processing has been explored in image
inpainting~\cite{chu2023rethinking} and crack segmentation under
shadows~\cite{zhang2024crack}.
Related work investigates high-frequency refinement for weakly
supervised segmentation~\cite{yang2025ffr} and frequency-aware
decoding for thin cracks~\cite{joo2026frequency}.
Directional modeling represents cracks as oriented
sub-cracks~\cite{chen2023devil}, while strip convolutions capture
elongated objects in remote sensing images~\cite{yuan2026strip}.
These studies motivate combining spectral and directional cues in
our grid classification branch.
Because cracks and pavement textures can share frequency components,
we combine input-dependent spectral weighting with directional refinement.

\subsection{Weak Supervision and Loss Reweighting}

Annotation requirements and foreground sparsity motivate alternative
supervision strategies.
Weakly supervised crack analysis uses coarse annotations or generated
labels to reduce dependence on precise masks
~\cite{he2024research,huang2023weakly,inoue2023weakly,konig2022weakly}.
Loss design has also been investigated for imbalanced crack
segmentation~\cite{liang2024lightweight,siriborvornratanakul2023pixel}.
Focal loss reduces the contribution of well-classified
examples~\cite{lin2017focal}.
Our supervision operates on category-specific grid occupancy.
Adaptive Importance Reweighting uses prediction uncertainty and local
grid-label transitions to weight focal loss while supervising all entries.
These transitions describe coarse occupancy changes without requiring
pixel-wise crack boundaries.

\begin{figure*}[t]
    \centering
    \includegraphics[width=\textwidth]{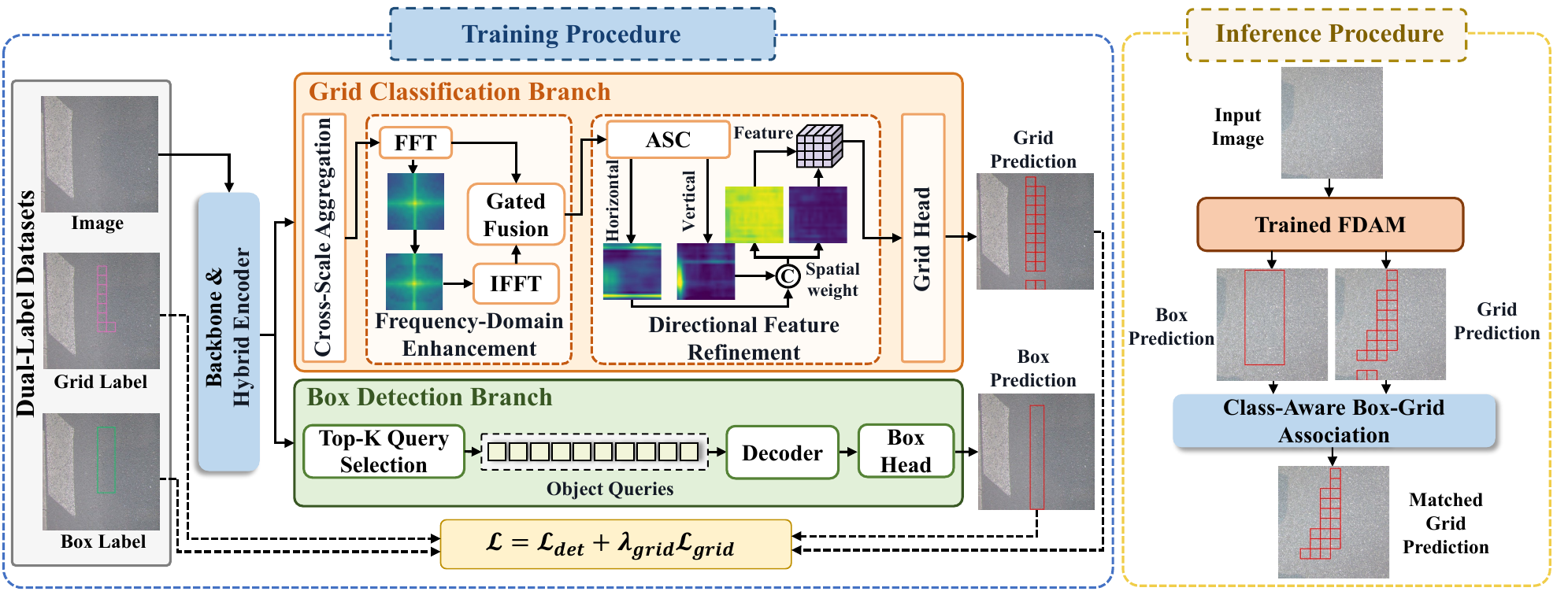}
    \caption{Overview of FDAM for pavement crack grid localization.
    The training procedure jointly optimizes a box detection branch
    and a grid classification branch through shared feature learning.
    The inference procedure applies class-aware box-grid association
    to refine grid predictions using detection categories and
    spatial overlap.}
    \label{fig:overall}
\end{figure*}

\section{Methodology}
\label{sec:method}

We formulate pavement crack grid localization as joint instance detection and category-wise grid occupancy prediction. FDAM couples shared feature learning with spectral enhancement, directional refinement, and adaptive grid supervision.

\subsection{Overall Framework}
\label{sec:overall}

Given an input image $\mathbf I\in\mathbb R^{3\times H\times W}$,
a backbone and hybrid encoder produce three shared feature maps
$\{\mathbf F_l\}_{l=1}^{3}$.
The two branches predict
\begin{equation}
    \mathcal D_{\mathrm{raw}}
    =\{(\hat{\mathbf p}_i,\hat{\mathbf b}_i)\}_{i=1}^{N_q},
    \quad \mathbf Z\in\mathbb R^{C_g\times H_g\times W_g},
    \label{eq:joint_outputs}
\end{equation}
where $N_q$ is the query count, and $\hat{\mathbf p}_i$ and
$\hat{\mathbf b}_i$ denote category scores and normalized boxes.
$\mathbf Z$ contains category-specific grid logits, where
$Z_c(u,v)$ predicts the presence of category $c$ in cell $(u,v)$
before sigmoid activation. The $C_g$ categories share an $H_g\times W_g$ grid.

As shown in Fig.~\ref{fig:overall}, both branches learn from shared
features while retaining task-specific prediction heads.
The detection branch follows DEIMv2~\cite{huang2025real}, using a
transformer decoder and Hungarian matching for set prediction.
We retain its classification, box regression, and auxiliary losses
in $\mathcal L_{\mathrm{det}}$. Training minimizes
\begin{equation}
    \mathcal L=\mathcal L_{\mathrm{det}}
    +\lambda_{\mathrm{grid}}\mathcal L_{\mathrm{grid}},
    \label{eq:total_loss}
\end{equation}
where $\lambda_{\mathrm{grid}}$ balances the objectives.
Both losses update the shared backbone and encoder.

\subsection{Grid Classification Branch}
\label{sec:grid_classification_branch}

\begin{figure*}[t]
    \centering
    \includegraphics[width=\textwidth]{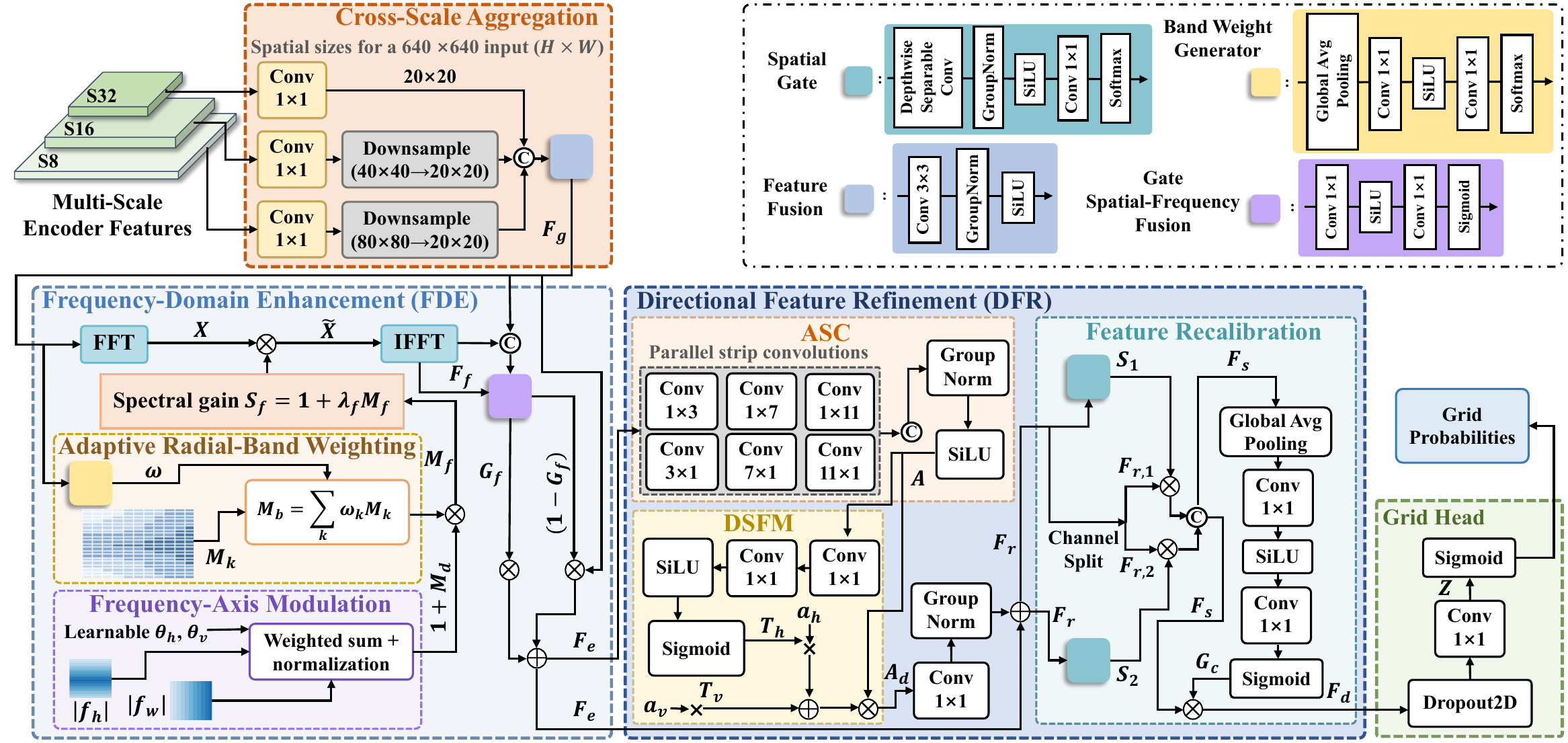}
    \caption{Grid classification branch with Frequency-Domain Enhancement
    and Directional Feature Refinement. Spatial dimensions illustrate
    a $640\times640$ input.}
    \label{fig:grid_classification_branch}
\end{figure*}

Pointwise projections and bilinear resizing align the shared features
to $C$ channels and the coarsest encoder resolution $(H_g,W_g)$.
Concatenation and convolution produce the fused feature $\mathbf F_g$.
Fig.~\ref{fig:grid_classification_branch} shows the subsequent
enhancement and prediction stages.
Detailed operators are given in Supplementary Sec.~\ref{supp:architecture}.

\subsubsection{Frequency-Domain Enhancement}
\label{sec:fde}

Thin and tiny cracks produce spatial variations across multiple
frequency bands, which may also contain pavement textures and noise.
FDE therefore learns input-dependent spectral weights.
Let $\mathcal F_r$ denote the orthonormal real Fourier transform.
Four disjoint radial masks $M_k$ receive weights
\begin{equation}
    \begin{aligned}
        \boldsymbol\omega
        &=\operatorname{Softmax}(\phi_b(\operatorname{GAP}(\mathbf F_g))),\\
        M_b&=\sum_{k=1}^{4}\omega_kM_k,
        M_f=M_b\odot(1+\chi_fM_d),
    \end{aligned}
    \label{eq:adaptive_band_mask}
\end{equation}
where $\operatorname{GAP}$ denotes global average pooling,
$M_d\in[0,1]$ is a learned axis-modulation mask, and
$\chi_f\in\{0,1\}$ specifies whether it is enabled.

The masks are shared across channels.
Spectral reconstruction and gated fusion give
\begin{equation}
    \begin{aligned}
        \mathbf F_f
        &=\mathcal F_r^{-1}\!\left[
        (1+\lambda_fM_f)\odot\mathcal F_r(\mathbf F_g)\right],\\
        \mathbf G_f&=\sigma(\phi_g(\mathbf F_g\Vert\mathbf F_f)),\\
        \mathbf F_{\mathrm e}
        &=\mathbf F_g+\mathbf G_f\odot(\mathbf F_f-\mathbf F_g),
    \end{aligned}
    \label{eq:fde_output}
\end{equation}
where $\lambda_f$ is learned, $\sigma$ is sigmoid,
$\Vert$ denotes channel concatenation, and $\odot$ denotes
element-wise multiplication.

The spatial gate is broadcast across channels.
Since $\|M_f\|_\infty\leq2$, the correction satisfies
\begin{equation}
    \|\mathbf F_{\mathrm e}-\mathbf F_g\|_F
    \leq2|\lambda_f|\,\|\mathbf F_g\|_F.
    \label{eq:fde_correction_bound}
\end{equation}

Thus, $|\lambda_f|$ controls the relative correction magnitude for
each input, and $\lambda_f=0$ recovers the input.
Mask construction and the derivation appear in
Supplementary Sec.~\ref{supp:spectral}.

\subsubsection{Directional Feature Refinement}
\label{sec:dfr}

DFR captures elongated crack patterns through multiscale strip
convolutions, directional modulation, and feature recalibration.
Given the input feature $\mathbf F_{\mathrm{in}}$, each DFR
stage applies parallel $1\times k$ and $k\times1$ convolutions
with $k\in\{3,7,11\}$, followed by concatenation, normalization,
and activation to obtain $\mathbf A$.
Axis-wise pooling produces attention maps $\mathbf T_h,\mathbf T_v$,
while global pooling predicts softmax weights $a_h,a_v$.
The directional update is
\begin{equation}
    \begin{aligned}
        \mathbf A_d
        &=\mathbf A\odot(a_h\mathbf T_h+a_v\mathbf T_v),\\
        \mathbf F_{\mathrm{out}}
        &=\mathbf F_{\mathrm{in}}
        +\operatorname{GN}(\operatorname{Conv}_{1\times1}(\mathbf A_d)).
    \end{aligned}
    \label{eq:dfr_residual}
\end{equation}
where $\operatorname{GN}$ denotes group normalization. The residual connection preserves the input features while incorporating the directional refinement.

Applying $N_b$ independently parameterized stages to
$\mathbf F_{\mathrm e}$ yields $\mathbf F_r$.
Two complementary spatial maps reweight equal channel groups,
followed by channel recalibration:
\begin{equation}
    \begin{aligned}
        \mathbf F_s
        &=(\mathbf F_{r,1}\odot\mathbf S_1)
        \Vert(\mathbf F_{r,2}\odot\mathbf S_2),\\
        \mathbf F_d
        &=\mathbf F_s\odot\sigma(\psi_c(\operatorname{GAP}(\mathbf F_s))).
    \end{aligned}
    \label{eq:dfr_output}
\end{equation}
where the maps satisfy $\mathbf S_1+\mathbf S_2=1$ at each location.
The channel groups are learned partitions without prescribed orientations.
Spatial dropout and a pointwise classifier produce $\mathbf Z$.
Supplementary Sec.~\ref{supp:directional} specifies the attention,
recalibration, and prediction operators.

\subsection{Adaptive Importance Reweighting}
\label{sec:air}

Grid supervision emphasizes uncertain predictions and local label
transitions while retaining supervision over every category-cell entry.
Grid annotations are stored separately from detection boxes.
Their transformed centers and eight offsets are projected onto the
prediction grid to form $\mathbf Y\in\{0,1\}^{C_g\times H_g\times W_g}$.
Categories are encoded independently, allowing multiple labels per cell.
Supplementary Sec.~\ref{supp:grid_targets} details target construction.

For training, let $\bar{\mathbf Z}=\operatorname{clip}(\mathbf Z,-20,20)$
and $\mathbf P=\sigma(\bar{\mathbf Z})$.
Define
\begin{equation}
    \begin{aligned}
        \mathbf U&=1-|2\mathbf P-1|,\\
        \mathbf T&=\operatorname{Dilate}_{k_t}(\mathbf Y)
                   -\operatorname{Erode}_{k_t}(\mathbf Y),\\
        \mathbf W_{\mathrm{AIR}}
        &=\operatorname{StopGrad}(1+\lambda_u\mathbf U+\lambda_t\mathbf T).
    \end{aligned}
    \label{eq:air_weight}
\end{equation}

Uncertainty peaks at $P=0.5$, and $\mathbf T$ marks coarse grid-label
transitions rather than pixel-wise crack boundaries.
StopGrad prevents differentiation through the weighting rule.
The grid objective uses binary focal loss~\cite{lin2017focal}:
\begin{equation}
    \mathcal L_{\mathrm{grid}}
    =\frac{\sum_j W_{\mathrm{AIR},j}
    \ell_{\mathrm{focal}}(P_j,Y_j)}{\max(1,N_+)},
     N_+=\sum_jY_j,
    \label{eq:air_loss}
\end{equation}
where $j=(b,c,u,v)$ indexes the local mini-batch.

With $\lambda_u=\lambda_t=1$, the weights satisfy $1\leq W_j\leq3$.
Relative to the unit-weight objective $\mathcal L_0$,
\begin{equation}
    \frac{\partial\mathcal L_{\mathrm{grid}}}{\partial\bar Z_j}
    =W_{\mathrm{AIR},j}
    \frac{\partial\mathcal L_0}{\partial\bar Z_j}.
    \label{eq:air_gradient}
\end{equation}

The weights preserve each clipped-logit's gradient sign while scaling
its magnitude by a bounded positive factor.
The focal formulation, hyperparameters, and derivation are provided
in Supplementary Sec.~\ref{supp:loss}.

\subsection{Inference Procedure}
\label{sec:inference}

A single forward pass produces both outputs.
Decoded detections with scores $s^d>\tau_d$ form $\mathcal D$.
Each grid cell selects its highest-scoring category from
$\sigma(\mathbf Z)$ and is retained as a candidate when $s^g\geq\tau_g$.
After mapping both outputs to original-image coordinates and a common
category space, association retains
\begin{equation}
    \mathcal G^*=\left\{g\in\mathcal G\;\middle|\;
    \begin{array}{l}
    \exists d\in\mathcal D:\ \hat c^g=\hat c^d,\\
    |\mathbf b^g\cap\mathbf b^d|>0
    \end{array}\right\}.
    \label{eq:refined_grid_set}
\end{equation}

Association only filters grid candidates and preserves their scores.
If $\mathcal D=\varnothing$, then $\mathcal G^*=\varnothing$.
Grid-only inference returns the unassociated set $\mathcal G$.
Box average precision uses decoded detections independently of
the association threshold.
Supplementary Sec.~\ref{supp:inference} and
Algorithm~\ref{alg:joint_inference} give the complete procedure.

\section{Experiments}
\label{sec:experiments}

We evaluate box detection and grid classification on RoadDistress2026,
Crack500, and DeepCrack, followed by component and task ablations. Additional evaluations stratified by instance extent and mask width, together with qualitative comparisons, are provided in Sec.~\ref{supp:additional_experiments} of the supplementary material.

\subsection{Datasets and Annotation Preparation}
\label{sec:datasets}

\noindent\textbf{RoadDistress2026.}
Our dataset contains 13,000 images at $4656\times2657$ pixels:
9,100 for training, 2,600 for validation, and 1,300 for testing.
Longitudinal cracks (LC) and transverse cracks (TC) are labeled on a
$20\times37$ grid. Positive cells carry category and instance labels;
each detection box encloses one instance's cells.

\noindent\textbf{Public Benchmarks.}
Crack500 contains 500 mask-labeled images \cite{yang2019feature},
processed into 1,896 training, 348 validation, and 1,124 test samples.
DeepCrack contains 537 images of diverse surfaces
\cite{zou2019deepcrack}, split into 375/53/109 images for
training/validation/testing.

For both datasets, boxes enclose 8-connected mask components with
at least three foreground pixels. Grid annotations are derived
separately using $20\times20$-pixel reference cells on a
$640\times640$ plane, scaled to the original image size and rounded
to integer pixels. These annotation cells differ from the network's
$20\times20$ output grid, whose cells cover $32\times32$ input pixels.
Annotation regions map to the output grid through the nine-point
projection in Section~\ref{sec:air}, with $\delta_g=0.125$.
FDAM uses the derived boxes and grid regions.
Segmentation baselines use dense masks.

\subsection{Experimental Setup}
\label{sec:implementation}

\subsubsection{Training Settings}

FDAM uses PyTorch 2.2.2 on an Intel Xeon E5-2680 workstation with
two 24~GB NVIDIA RTX 3090 GPUs. AdamW uses $\beta_1=0.9$,
$\beta_2=0.999$, warm-up, and a flat-cosine schedule.
Backbones use pretrained DINOv3 representations. DeepCrack and its
ablations share COCO-pretrained DEIMv2 initialization and seed zero.
Training applies ImageNet normalization, photometric and geometric
augmentation, scheduled Mosaic/MixUp, and multiscale resizing.
The grid head uses four frequency bands and dropout 0.1. Grid loss
parameters are $\alpha_f=0.5$ and $\gamma_f=1.5$. AIR uses
$\lambda_u=\lambda_t=1$ and $k_t=3$.

\begin{table*}[!tp]
    \centering
    \caption{Box detection on Crack500 and DeepCrack.
    All methods use the same mask-derived box annotations.
    Best and second-best metric values are \bestval{bold}
    and \secondval{underlined}, respectively.}
    \label{tab:public_detection}
    \small
    \setlength{\tabcolsep}{2pt}
    \renewcommand{\arraystretch}{1.12}
    \begin{tabularx}{\textwidth}{@{}l*{10}{C}@{}}
        \toprule
        \rowcolor{FDAMHeader}
        & \multicolumn{5}{c}{Crack500} & \multicolumn{5}{c}{DeepCrack} \\
        \cmidrule(lr){2-6}\cmidrule(lr){7-11}
        \rowcolor{FDAMHeader}
        \multirow{-2}{*}{Method}
        & \shortstack{Params\\(M)} & GFLOPs & AP & AP$_{50}$ & AP$_{75}$
        & \shortstack{Params\\(M)} & GFLOPs & AP & AP$_{50}$ & AP$_{75}$ \\
        \midrule
        YOLOv12~\cite{tian2026yolov12} & 26 & 99 & 18.8 & 33.8 & 18.2 & 26 & 99 & 32.8 & 52.0 & 32.6 \\
        YOLO26~\cite{sapkota2025yolo26} & 25 & 96 & 18.0 & 33.1 & 17.1 & 25 & 96 & 33.2 & 52.3 & 36.3 \\
        RT-DETR~\cite{zhao2024detrs} & 36 & 100 & 19.8 & 34.1 & 19.6 & 36 & 100 & 35.8 & 52.0 & 36.3 \\
        RT-DETRv2~\cite{lv2024rt} & 36 & 100 & 19.9 & 33.9 & 20.0 & 36 & 100 & 30.2 & 44.7 & 30.6 \\
        RT-DETRv3~\cite{wang2025rt} & 31 & 92 & 18.6 & 31.7 & 18.5 & 31 & 92 & 26.7 & 39.5 & 28.7 \\
        RT-DETRv4~\cite{liao2025rt} & 31 & 91 & 20.1 & 34.8 & 20.0 & 31 & 91 & 25.6 & 40.6 & 24.5 \\
        DEIM~\cite{huang2025deim} & 31 & 92 & 20.4 & 35.0 & 20.3 & 31 & 92 & 36.9 & 54.7 & \secondval{37.7} \\
        DEIMv2~\cite{huang2025real} & 10 & 25 & \secondval{21.3} & \secondval{35.4} & \secondval{21.0} & 18 & 52 & \secondval{37.6} & \secondval{55.5} & 37.3 \\
        \midrule
        \textbf{FDAM (Ours)} & 12 & 29 & \bestval{21.8} & \bestval{36.6} & \bestval{21.8} & 22 & 56 & \bestval{39.8} & \bestval{57.3} & \bestval{40.7} \\
        \bottomrule
    \end{tabularx}
\end{table*}

\begin{table*}[!tp]
    \centering
    \caption{Box detection on RoadDistress2026.
    FDAM uses only the detection branch in this comparison.
    LC and TC denote longitudinal and transverse cracks.
    Best and second-best metric values are \bestval{bold}
    and \secondval{underlined}, respectively.}
    \label{tab:road_detection}
    \small
    \setlength{\tabcolsep}{2pt}
    \renewcommand{\arraystretch}{1.12}
    \begin{tabularx}{\textwidth}{@{}lcc*{9}{C}@{}}
        \toprule
        \rowcolor{FDAMHeader}
        & & & \multicolumn{3}{c}{LC}
        & \multicolumn{3}{c}{TC} & \multicolumn{3}{c}{All} \\
        \cmidrule(lr){4-6}\cmidrule(lr){7-9}\cmidrule(lr){10-12}
        \rowcolor{FDAMHeader}
        \multirow{-2}{*}{Method}
        & \multirow{-2}{*}{\shortstack{Params\\(M)}}
        & \multirow{-2}{*}{GFLOPs}
        & AP & AP$_{50}$ & AP$_{75}$
        & AP & AP$_{50}$ & AP$_{75}$
        & AP & AP$_{50}$ & AP$_{75}$ \\
        \midrule
        YOLOv12~\cite{tian2026yolov12} & 26 & 281 & 15.6 & 38.5 & 8.8 & 16.5 & 40.4 & 11.0 & 16.1 & 39.5 & 9.9 \\
        YOLO26~\cite{sapkota2025yolo26} & 25 & 248 & 15.1 & 34.4 & 11.1 & 15.9 & 38.1 & 10.5 & 15.5 & 36.3 & 10.8 \\
        RT-DETR~\cite{zhao2024detrs} & 36 & 254 & 15.4 & 38.7 & 8.8 & 15.9 & 42.9 & 7.2 & 15.7 & 40.8 & 8.0 \\
        RT-DETRv2~\cite{lv2024rt} & 36 & 254 & 15.0 & 37.1 & 10.3 & 16.5 & 42.8 & 9.1 & 15.8 & 40.0 & 9.7 \\
        RT-DETRv3~\cite{wang2025rt} & 43 & 157 & 15.5 & 39.1 & 8.8 & 15.3 & 39.2 & 9.8 & 15.4 & 39.2 & 9.3 \\
        RT-DETRv4~\cite{liao2025rt} & 31 & 224 & \secondval{19.9} & \secondval{50.2} & 12.7 & \secondval{19.1} & \secondval{50.0} & 11.2 & \secondval{19.5} & \secondval{50.1} & 12.0 \\
        DEIM~\cite{huang2025deim} & 31 & 230 & 19.6 & 47.4 & \secondval{12.8} & 18.6 & 47.2 & 11.4 & 19.1 & 47.3 & 12.1 \\
        DEIMv2~\cite{huang2025real} & 32 & 242 & \secondval{19.9} & 48.8 & 12.6 & \secondval{19.1} & \secondval{50.0} & \secondval{11.7} & \secondval{19.5} & 49.4 & \secondval{12.2} \\
        \midrule
        \textbf{FDAM (Ours)} & 35 & 248 & \bestval{20.2} & \bestval{51.1} & \bestval{13.5} & \bestval{19.8} & \bestval{51.1} & \bestval{12.2} & \bestval{20.0} & \bestval{51.1} & \bestval{12.9} \\
        \bottomrule
    \end{tabularx}
\end{table*}

\subsubsection{Model Selection and Inference}

\begin{figure*}[!tp]
    \centering
    \includegraphics[width=\textwidth]{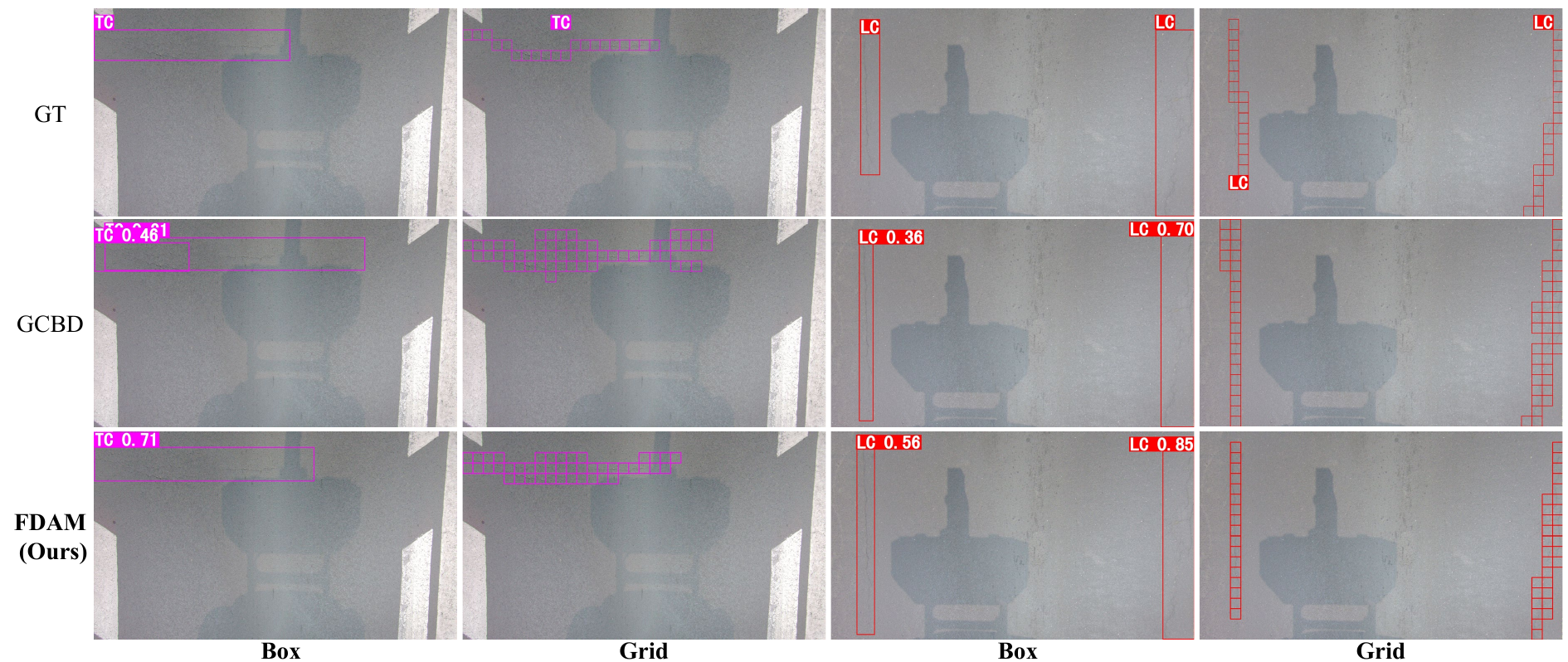}
    \caption{Qualitative comparison of GCBD and the joint FDAM model on
    RoadDistress2026 for box detection and grid localization.}
    \label{fig:multi_comparison}
\end{figure*}

\begin{table*}[!tp]
    \centering
    \caption{Joint box--grid comparison on RoadDistress2026.
    FDAM uses the multitask model with both detection and grid branches.
    All denotes the average over LC and TC. Grid AP is class-wise
    average precision without an IoU sweep.
    Best and second-best metric values are \bestval{bold}
    and \secondval{underlined}, respectively.}
    \label{tab:multitask_comparison}
    \label{tab:multitask_box}
    \label{tab:multitask_grid}
    \small
    \setlength{\tabcolsep}{2pt}
    \renewcommand{\arraystretch}{1.12}
    \begin{tabularx}{\textwidth}{@{}llcc*{9}{C}@{}}
        \toprule
        \rowcolor{FDAMHeader}
        & & & & \multicolumn{3}{c}{LC}
        & \multicolumn{3}{c}{TC} & \multicolumn{3}{c}{All} \\
        \cmidrule(lr){5-7}\cmidrule(lr){8-10}\cmidrule(lr){11-13}
        \rowcolor{FDAMHeader}
        \multirow{-2}{*}{Task} & \multirow{-2}{*}{Method}
        & \multirow{-2}{*}{\shortstack{Params\\(M)}}
        & \multirow{-2}{*}{GFLOPs}
        & AP & AP$_{50}$ & AP$_{75}$
        & AP & AP$_{50}$ & AP$_{75}$
        & AP & AP$_{50}$ & AP$_{75}$ \\
        \midrule
        \multirow{2}{*}{Box} & GCBD~\cite{li2023grid} & 56 & 117 & \secondval{7.6} & \secondval{25.5} & \secondval{4.0} & \secondval{13.4} & \secondval{41.8} & \secondval{3.6} & \secondval{10.5} & \secondval{33.7} & \secondval{3.8} \\
         & \textbf{FDAM (Ours)} & 35 & 248 & \bestval{17.7} & \bestval{44.5} & \bestval{12.4} & \bestval{19.6} & \bestval{46.3} & \bestval{13.0} & \bestval{18.7} & \bestval{45.4} & \bestval{12.7} \\
        \midrule
        \rowcolor{FDAMHeader}
        & & & & P & R & AP & P & R & AP & P & R & AP \\
        \midrule
        \multirow{2}{*}{Grid} & GCBD~\cite{li2023grid} & 56 & 117 & \secondval{44.0} & \secondval{61.6} & \secondval{46.7} & \secondval{53.5} & \secondval{55.6} & \secondval{51.3} & \secondval{48.8} & \secondval{58.6} & \secondval{49.0} \\
         & \textbf{FDAM (Ours)} & 35 & 248 & \bestval{59.3} & \bestval{68.8} & \bestval{65.6} & \bestval{62.0} & \bestval{69.1} & \bestval{67.3} & \bestval{60.7} & \bestval{69.0} & \bestval{66.5} \\
        \bottomrule
    \end{tabularx}
\end{table*}

\begin{table*}[!tp]
    \centering
    \caption{Grid classification results for pavement crack grid
    localization on Crack500 and DeepCrack under a common grid
    protocol. Overall metrics are macro-averaged over background and
    crack. Segmentation baselines use dense mask supervision, while
    FDAM uses derived box and grid annotations.
    Params and GFLOPs are given as Crack500/DeepCrack when they differ.
    Best and second-best metric values are \bestval{bold}
    and \secondval{underlined}, respectively.}
    \label{tab:segmentation_comparison}
    \footnotesize
    \setlength{\tabcolsep}{1pt}
    \renewcommand{\arraystretch}{1.12}
    \begin{tabularx}{\textwidth}{@{}lcc*{16}{C}@{}}
        \toprule
        \rowcolor{FDAMHeader}
        & & & \multicolumn{8}{c}{Crack500}
        & \multicolumn{8}{c}{DeepCrack} \\
        \cmidrule(lr){4-11}\cmidrule(lr){12-19}
        \rowcolor{FDAMHeader}
        & & & \multicolumn{4}{c}{Overall}
        & \multicolumn{4}{c}{Crack Only}
        & \multicolumn{4}{c}{Overall}
        & \multicolumn{4}{c}{Crack Only} \\
        \cmidrule(lr){4-7}\cmidrule(lr){8-11}
        \cmidrule(lr){12-15}\cmidrule(lr){16-19}
        \rowcolor{FDAMHeader}
        \multirow{-3}{*}{Method}
        & \multirow{-3}{*}{\shortstack{Params\\(M)}}
        & \multirow{-3}{*}{GFLOPs}
        & mP & mR & mF1 & mIoU & P & R & F1 & IoU
        & mP & mR & mF1 & mIoU & P & R & F1 & IoU \\
        \midrule
        SimCrack~\cite{jaziri2024designing} & 299 & 146 & 84.3 & 86.2 & 85.2 & 75.7 & 72.3 & 77.2 & 74.7 & 59.6 & 89.0 & 86.5 & 87.7 & 79.2 & 82.4 & 76.0 & 79.0 & 65.3 \\
        SCSegamba~\cite{liu2025scsegamba} & 2.8 & 18 & \secondval{90.5} & 90.6 & 90.5 & 83.4 & \secondval{83.5} & 83.8 & 83.7 & 71.9 & \secondval{94.2} & 89.6 & 91.7 & 85.2 & \secondval{91.8} & 80.5 & 85.8 & 75.1 \\
        MixerCSeg~\cite{zhao2026mixercseg} & 2.5 & 3 & 89.9 & 90.3 & 90.1 & 82.8 & 82.4 & 83.5 & 82.9 & 70.9 & 90.6 & 89.0 & 89.8 & 82.2 & 84.7 & 80.7 & 82.6 & 70.4 \\
        CCMamba~\cite{liu2026ccmamba} & 89 & 41 & 90.3 & \secondval{91.4} & \secondval{90.8} & \secondval{83.9} & 82.9 & \secondval{85.7} & \secondval{84.3} & \secondval{72.8} & \bestval{95.7} & \secondval{91.3} & \secondval{93.3} & \secondval{87.8} & \bestval{94.4} & \secondval{83.4} & \secondval{88.6} & \secondval{79.5} \\
        MSFAF-Net~\cite{li2026msfaf} & 5.5 & 28 & 87.0 & 88.6 & 87.8 & 79.4 & 77.1 & 81.1 & 79.1 & 65.4 & 87.5 & 85.5 & 86.4 & 77.3 & 79.5 & 74.4 & 76.9 & 62.4 \\
        \midrule
        \textbf{FDAM (Ours)} & 12/22 & 29/56 & \bestval{91.5} & \bestval{92.7} & \bestval{92.1} & \bestval{85.8} & \bestval{84.9} & \bestval{87.8} & \bestval{86.3} & \bestval{76.0} & 93.7 & \bestval{93.7} & \bestval{93.7} & \bestval{88.5} & 89.3 & \bestval{89.4} & \bestval{89.4} & \bestval{80.8} \\
        \bottomrule
    \end{tabularx}
\end{table*}

Validation-selected checkpoints and thresholds are frozen for testing
with exponential-moving-average weights. Joint and detection-only
models are selected by box AP. Grid-only uses foreground grid F1.
Both joint outputs use the same checkpoint. On public datasets,
thresholds maximize validation foreground grid F1, including for
segmentation baselines. Joint models use $\tau_g=0.45$ and
$\tau_d=0.30$. DeepCrack Grid-only uses $\tau_g=0.50$.

Joint inference retains grid cells overlapping same-category
detections with positive area. Grid-only uses raw grids. Box AP is
evaluated independently of grid association. RoadDistress2026 uses
$\tau_d=\tau_g=0.5$ for quantitative joint inference.
All three datasets retain grid candidates with $s^g\geq\tau_g$
and detection candidates with $s^d>\tau_d$.

\subsubsection{Evaluation Protocol}

Comparisons include eight box detectors, five segmentation methods,
and GCBD~\cite{li2023grid} under identical splits. Box detection uses
COCO AP over IoU thresholds 0.50:0.05:0.95, $\mathrm{AP}_{50}$, and
$\mathrm{AP}_{75}$. RoadDistress2026 reports category-wise grid
precision, recall, and confidence-ranked cell AP after box association,
without an IoU sweep. On public datasets, FDAM probabilities are
thresholded directly on its output grid. Precision, recall, F1, and
IoU use the confusion matrix aggregated over all test cells. Overall
metrics average background and crack equally. Crack Only denotes
foreground metrics. All accuracy values are percentages.


\subsubsection{Input Resolution and Complexity}

The public-dataset comparisons in Table~\ref{tab:public_detection}
use $640\times640$ inputs. RoadDistress2026 comparisons in
Tables~\ref{tab:road_detection} and~\ref{tab:multitask_comparison}
use $1024\times1024$ inputs. In
Table~\ref{tab:segmentation_comparison}, GFLOPs use each method's input
resolution: $256\times256$ for SimCrack and MSFAF-Net,
$512\times512$ for SCSegamba and MixerCSeg, and $640\times640$ for
FDAM. CCMamba's GFLOPs cover one $384\times384$ sliding window,
excluding additional windows per image. 

\subsection{Comparison with Existing Methods}
\label{sec:sota_comparison}

\subsubsection{Box Detection on Public Benchmarks}

Table~\ref{tab:public_detection} shows that FDAM achieves
21.8/39.8 AP on Crack500/DeepCrack, exceeding the
strongest baselines by 0.5/2.2 percentage points and leading in
$\mathrm{AP}_{50}$ and $\mathrm{AP}_{75}$.

\subsubsection{Box Detection on RoadDistress2026}

In Table~\ref{tab:road_detection}, the detection-only configuration
reaches 20.0 AP, 51.1 $\mathrm{AP}_{50}$, and 12.9
$\mathrm{AP}_{75}$, leading for both LC and TC.
The joint model in Table~\ref{tab:multitask_comparison}
performs both box detection and grid classification, achieving
18.7 box AP and 66.5 grid AP. Fig.~\ref{fig:multi_comparison} presents qualitative comparisons of the detection and grid-classification results between
the two methods.

\subsubsection{Comparison on Pavement Crack Grid Localization}

\begin{figure*}[!tp]
    \centering
    \includegraphics[width=\textwidth]{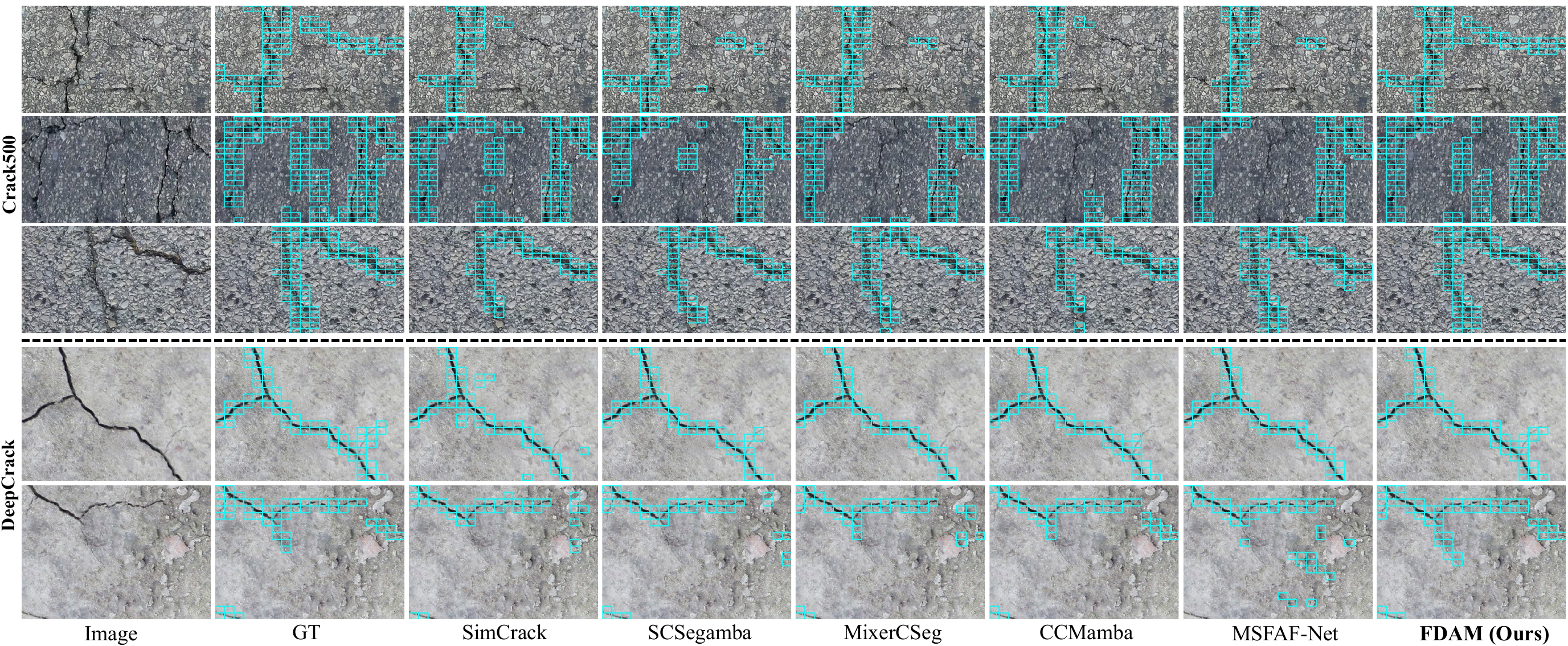}
    \caption{Qualitative comparison of grid localization on
    Crack500 and DeepCrack. Each row shows an input image,
    grid ground truth, and predictions from the compared methods.
    Segmentation outputs are converted to the common grid
    representation.}
    \label{fig:grid_qualitative_comparison}
\end{figure*}

Table~\ref{tab:segmentation_comparison} compares FDAM with segmentation
methods after converting their predictions to grids. FDAM leads all eight
metrics on Crack500, with mF1/mIoU of 92.1/85.8, exceeding CCMamba
by 1.3/1.9 percentage points. On DeepCrack, FDAM achieves the highest
mF1/mIoU of 93.7/88.5. Relative to CCMamba, higher crack recall
(89.4 versus 83.4) offsets lower precision (89.3 versus 94.4),
yielding higher crack F1/IoU of 89.4/80.8.
Fig.~\ref{fig:grid_qualitative_comparison} shows qualitative comparisons
under the same grid protocol.

\subsection{Ablation Studies}
\label{sec:ablation}

DeepCrack ablations share the backbone, split, initialization, and
training settings except for the component or task removed.
Both joint outputs use one validation-selected checkpoint.
Table~\ref{tab:ablation_main} reports task and component ablations
with all detection and grid metrics.

\begin{table}[!htbp]
    \centering
    \caption{Task and component ablations on DeepCrack.
    Grid metrics average background and crack.
    Best and second-best metric values are \bestval{bold}
    and \secondval{underlined}, respectively.}
    \label{tab:ablation_main}
    \label{tab:ablation_components}
    \footnotesize
    \setlength{\tabcolsep}{1pt}
    \renewcommand{\arraystretch}{1.12}
    \begin{tabularx}{\columnwidth}{@{}l*{7}{C}@{}}
        \toprule
        \rowcolor{FDAMHeader}
        & \multicolumn{3}{c}{Box Detection}
        & \multicolumn{4}{c}{Grid Classification} \\
        \cmidrule(lr){2-4}\cmidrule(lr){5-8}
        \rowcolor{FDAMHeader}
        \multirow{-2}{*}{Configuration}
        & AP & AP$_{50}$ & AP$_{75}$ & mP & mR & mF1 & mIoU \\
        \midrule
        Detection-only & 35.5 & 51.1 & 36.8 & -- & -- & -- & -- \\
        Grid-only & -- & -- & -- & \secondval{93.8} & \bestval{94.5} & \bestval{94.2} & \bestval{89.2} \\
        \midrule
        Joint w/o FDE & 38.4 & 54.7 & \secondval{40.4} & 93.3 & 93.6 & 93.4 & 88.0 \\
        Joint w/o DFR & \secondval{38.6} & 54.1 & 38.7 & 93.3 & 93.4 & 93.4 & 87.9 \\
        Joint w/o AIR & 37.8 & \secondval{54.8} & 38.7 & \bestval{95.6} & 87.4 & 90.9 & 84.0 \\
        \midrule
        \textbf{Full FDAM} & \bestval{39.8} & \bestval{57.3} & \bestval{40.7} & 93.7 & \secondval{93.7} & \secondval{93.7} & \secondval{88.5} \\
        \bottomrule
    \end{tabularx}
\end{table}

\noindent\textbf{Adaptive Importance Reweighting.}
AIR increases box AP from 37.8 to 39.8 and grid mIoU from 84.0 to 88.5.
Higher grid macro recall, 93.7 versus 87.4, offsets lower macro
precision and raises mF1 from 90.9 to 93.7.
AIR is enabled in the grid-only and full joint models.

\noindent\textbf{Task branches.}
The joint model improves detection AP from 35.5 to 39.8
over detection-only. This gain suggests that grid supervision benefits detection through shared features, while box predictions provide category
and spatial constraints for grid predictions during inference, highlighting their complementary roles. Grid-only achieves higher mF1/mIoU, 94.2/89.2 versus 93.7/88.5. However, this comparison does not isolate a multitask trade-off. Joint models use box-AP checkpoint selection and associated grids,
whereas grid-only uses foreground-F1 selection and raw grids.

\noindent\textbf{Feature enhancement.}
Removing FDE reduces box AP/grid mIoU from 39.8/88.5 to 38.4/88.0.
Removing DFR reduces them to 38.6/87.9.
Both components improve the measured detection and grid scores
under these settings.


\subsubsection{Effect of the Grid Loss Weight}
\label{sec:grid_loss_weight}

Table~\ref{tab:grid_loss_weight} compares
$\lambda_{\mathrm{grid}}\in\{0.5,1.0,2.0\}$ on DeepCrack with other
training settings fixed. Each setting uses a checkpoint selected by
validation box AP for both tasks and validation-selected grid
thresholds frozen for testing. The setting $\lambda_{\mathrm{grid}}=1.0$ achieves the highest box AP and grid mIoU among the tested values.
Grid mF1 varies from 93.5 to 93.7 across these settings.

\begin{table}[!htbp]
    \centering
    \caption{Effect of the grid loss weight on DeepCrack.
    Grid metrics are macro-averaged over background and crack.
    All results are percentages.
    Best and second-best metric values are \bestval{bold}
    and \secondval{underlined}, respectively.}
    \label{tab:grid_loss_weight}
    \footnotesize
    \setlength{\tabcolsep}{1pt}
    \renewcommand{\arraystretch}{1.12}
    \begin{tabularx}{\columnwidth}{@{}c*{7}{C}@{}}
        \toprule
        \rowcolor{FDAMHeader}
        & \multicolumn{3}{c}{Box Detection}
        & \multicolumn{4}{c}{Grid Localization} \\
        \cmidrule(lr){2-4}\cmidrule(lr){5-8}
        \rowcolor{FDAMHeader}
        \multirow{-2}{*}{$\lambda_{\mathrm{grid}}$}
        & AP & AP$_{50}$ & AP$_{75}$ & mP & mR & mF1 & mIoU \\
        \midrule
        0.5 & 37.7 & \secondval{55.2} & 38.1 & \secondval{93.5} & 93.5 & \secondval{93.5} & 88.1 \\
        1.0 & \bestval{39.8} & \bestval{57.3} & \bestval{40.7} & \bestval{93.7} & \secondval{93.7} & \bestval{93.7} & \bestval{88.5} \\
        2.0 & \secondval{38.2} & 54.7 & \secondval{38.6} & 93.4 & \bestval{93.9} & \bestval{93.7} & \secondval{88.4} \\
        \bottomrule
    \end{tabularx}
\end{table}

\section{Conclusion}
\label{sec:conclusion}

We present FDAM for pavement crack grid localization of thin
and tiny cracks. The framework couples box detection and grid classification
through shared features. Spectral and directional enhancement strengthen grid representations, while adaptive importance reweighting emphasizes uncertain
predictions and label transitions. Experiments on RoadDistress2026, Crack500, and DeepCrack demonstrate the effectiveness of the framework and its components. Fixed grids limit geometric detail, and missed detections can
suppress valid cells during box-grid association.
\section*{LLM Usage Statement}
We used ChatGPT for language polishing and grammatical correction. All technical content and conclusions were developed and verified solely by the authors.
\section*{Acknowledgment}
The work was supported by the National Key Research and Development Program of China under Grant 2022YFA1003800, the National Natural Science Foundation of China under the Grant 62571273, Grant 62125102 and Grant U24B20177, and the Natural Science Foundation of Tianjin under the Grant 25JCLMJC01090.


{
    \small
    \bibliographystyle{ieeenat_fullname}
    \bibliography{main} 
}

\clearpage
\setcounter{page}{1}
\renewcommand{\thepage}{S\arabic{page}}
\maketitlesupplementary

\appendix
\numberwithin{equation}{section}

\section{Detailed Architecture}
\label{supp:architecture}

\subsection{Feature Aggregation and Detection}
\label{supp:aggregation}

The shared encoder produces feature maps
$\mathbf F_l\in\mathbb R^{C_l\times H_l\times W_l}$ at three scales.
For channel width $C$ divisible by $32$, pointwise projections $\phi_l$
and bilinear interpolation align them to $(H_g,W_g)=(H_3,W_3)$:
\begin{equation}
    \begin{aligned}
        \mathbf Q_l
        &=\operatorname{Resize}_{H_g,W_g}(\phi_l(\mathbf F_l)),\\
        \mathbf F_g
        &=\delta\!\left(\operatorname{GN}\!\left[
        \operatorname{Conv}_{3\times3}
        (\mathbf Q_1\Vert\mathbf Q_2\Vert\mathbf Q_3)\right]\right).
    \end{aligned}
    \label{eq:multiscale_fusion}
\end{equation}
where $\delta$ is SiLU.

The detection decoder progressively updates $N_q$ queries from
the shared encoder features. Prediction heads produce class scores and normalized boxes. Hungarian matching establishes one-to-one assignments to ground-truth
instances, and the original detection losses are retained.

\subsection{Directional Refinement and Prediction}
\label{supp:directional}

For each refinement stage, six convolution branches use
$\mathcal K=\{3,7,11\}$:
\begin{equation}
    \begin{aligned}
        \mathbf A_h
        &=\mathop{\Vert}_{k\in\mathcal K}
        \operatorname{Conv}_{1\times k}(\mathbf F_{\mathrm{in}}),\\
        \mathbf A_v
        &=\mathop{\Vert}_{k\in\mathcal K}
        \operatorname{Conv}_{k\times1}(\mathbf F_{\mathrm{in}}),\\
        \mathbf A
        &=\delta(\operatorname{GN}(\mathbf A_h\Vert\mathbf A_v)).
    \end{aligned}
    \label{eq:strip_aggregation}
\end{equation}

The six branches jointly output $C$ channels.
Each receives $\lfloor C/6\rfloor$ channels, with the remainder
assigned to the earlier branches in concatenation order.
Symmetric padding preserves spatial dimensions.

Direction-sensitive feature modulation uses
\begin{equation}
    \begin{aligned}
        \mathbf T_h
        &=\sigma(\phi_h(\operatorname{Avg}_H(\mathbf A))),\\
        \mathbf T_v
        &=\sigma(\phi_v(\operatorname{Avg}_W(\mathbf A))),\\
        [a_h,a_v]
        &=\operatorname{Softmax}(\phi_d(\operatorname{GAP}(\mathbf A))).
    \end{aligned}
    \label{eq:dsfm_attention}
\end{equation}

The maps have shapes $C\times1\times W_g$ and $C\times H_g\times1$
and are broadcast over the missing spatial dimension.
Each of $\phi_h,\phi_v$ consists of pointwise projections
$C\rightarrow\max(\lfloor C/4\rfloor,32)\rightarrow C$,
with SiLU between them. The mapping $\phi_d$ is a pointwise projection from $C$ to two outputs.

The resulting maps enter Eq.~\ref{eq:dfr_residual}.
Because $a_h+a_v=1$ and the attention maps lie in $(0,1)$,
$|\mathbf A_d|\leq|\mathbf A|$ element-wise before residual projection.
This property does not bound the entire residual block.

After $N_b$ stages, spatial recalibration predicts
\begin{equation}
    [\mathbf S_1,\mathbf S_2]
    =\operatorname{Softmax}(\psi_s(\mathbf F_r)).
    \label{eq:dfr_spatial_weights}
\end{equation}

The mapping $\psi_s$ applies a depthwise $3\times3$ convolution,
a pointwise projection to $C/4$ channels, eight-group normalization,
SiLU, and a two-output pointwise projection. The channel mapping $\psi_c$ in Eq.~\ref{eq:dfr_output} uses
pointwise projections $C\rightarrow C/4\rightarrow C$ with SiLU.

All spatial and channel gates are broadcast over the corresponding
dimensions when multiplied with features.
The final head computes
\begin{equation}
    \mathbf Z=\operatorname{Conv}_{1\times1}^{\mathrm{cls}}
    (\operatorname{Dropout2D}(\mathbf F_d)),
     \mathbf P=\sigma(\mathbf Z).
    \label{eq:grid_prediction}
\end{equation}

Independent sigmoid probabilities permit multiple positive categories
per training cell. Inference retains only the highest-scoring category.

\section{Spectral Masks and Correction Bound}
\label{supp:spectral}

\subsection{Frequency Coordinates and Masks}
\label{supp:spectral_masks}

The orthonormal real Fourier transform stores
$\mathbf X=\mathcal F_r(\mathbf F_g)$ with
$W_f=\lfloor W_g/2\rfloor+1$ columns.
Indices satisfy $0\leq\xi<H_g$ and $0\leq\eta<W_f$.
Mask coordinates follow the implementation:
\begin{equation}
    \begin{gathered}
        f_h(\xi)=\frac{\tilde\xi}{H_g},
        f_w(\eta)=\frac{\eta}{2W_f-1},\\
        \tilde\xi=
        \begin{cases}
            \xi,&\xi\leq\lfloor H_g/2\rfloor,\\
            \xi-H_g,&\text{otherwise}.
        \end{cases}
    \end{gathered}
    \label{eq:frequency_coordinates}
\end{equation}

These coordinates construct masks and do not redefine the Fourier
transform frequencies. In particular, the width denominator is
$W_g+1$ for even $W_g$ and $W_g$ for odd $W_g$.
Writing $q(\xi,\eta)=\sqrt{f_h(\xi)^2+f_w(\eta)^2}$ gives
\begin{equation}
    \begin{aligned}
        r(\xi,\eta)
        &=\frac{q(\xi,\eta)}
        {\max_{\xi',\eta'}q(\xi',\eta')+\epsilon},\\
        M_k
        &=\mathds{1}\!\left[\frac{k-1}{4}\leq r<\frac{k}{4}\right].
    \end{aligned}
    \label{eq:frequency_band}
\end{equation}

The four radial intervals are equally spaced and disjoint.
All maxima cover retained frequency bins, and $\epsilon=10^{-8}$.
The band predictor $\phi_b$ uses pointwise projections
$C\rightarrow\max(\lfloor C/4\rfloor,16)\rightarrow4$ with SiLU.
Softmax makes the weights positive and sum to one, hence
$0\leq M_b\leq1$.

When frequency-axis modulation is enabled, its mask is
\begin{equation}
    \begin{aligned}
        \relax[a_h^f,a_v^f]
        &=\operatorname{Softmax}([\theta_h,\theta_v]),\\
        D(\xi,\eta)
        &=a_h^f|f_h(\xi)|+a_v^f|f_w(\eta)|,\\
        M_d(\xi,\eta)
        &=\frac{D(\xi,\eta)}
        {\max_{\xi',\eta'}D(\xi',\eta')+\epsilon}.
    \end{aligned}
    \label{eq:direction_frequency_mask}
\end{equation}

The learned scalars $\theta_h,\theta_v$ are initialized to $0.5$
and shared across inputs and channels.
The configuration flag $\chi_f$ in Eq.~\ref{eq:adaptive_band_mask}
enables or disables this term.
The spectral scale $\lambda_f$ is unconstrained and initialized to $0.1$.
The fusion mapping $\phi_g$ uses pointwise projections
$2C\rightarrow\max(\lfloor C/4\rfloor,16)\rightarrow1$ with SiLU.
Its sigmoid output is a spatial gate shared across channels.
The real gain preserves a nonzero coefficient's phase wherever
$1+\lambda_fM_f>0$.

The implementation evaluates the orthonormal transform using real
cosine and sine matrices, including the conjugate-frequency
multiplicities required by the inverse transform.

\subsection{Bound on the Spectral Correction}
\label{supp:spectral_bound}

Let $\mathcal F$ denote the orthonormal full complex Fourier transform,
and let $\widehat M_f$ be the conjugate-symmetric extension of $M_f$.
The radial and absolute-axis masks preserve the required symmetry.
The reconstruction residual is
\begin{equation}
    \mathbf R=\mathbf F_f-\mathbf F_g
    =\lambda_f\mathcal F^{-1}
    (\widehat M_f\odot\mathcal F(\mathbf F_g)).
\end{equation}

Parseval's identity and $\|\widehat M_f\|_\infty\leq1+\chi_f$
imply
\begin{equation}
    \begin{aligned}
        \|\mathbf R\|_F
        &=|\lambda_f|\,
        \|\widehat M_f\odot\mathcal F(\mathbf F_g)\|_F\\
        &\leq(1+\chi_f)|\lambda_f|\,\|\mathbf F_g\|_F.
    \end{aligned}
\end{equation}

Since every gate value lies in $(0,1)$,
\begin{equation}
    \|\mathbf F_{\mathrm e}-\mathbf F_g\|_F
    =\|\mathbf G_f\odot\mathbf R\|_F
    \leq\|\mathbf R\|_F
    \leq2|\lambda_f|\,\|\mathbf F_g\|_F.
\end{equation}

Disabling axis modulation gives the tighter factor $|\lambda_f|$.
The argument uses the full spectrum, avoiding an unweighted half-spectrum interpretation of Parseval's identity. This is a per-input correction bound rather than a bound on sensitivity to input perturbations.

\section{Grid Supervision}
\label{supp:supervision}

\subsection{Grid Target Construction}
\label{supp:grid_targets}

\begin{figure}[t]
    \centering
    \includegraphics[width=\linewidth]{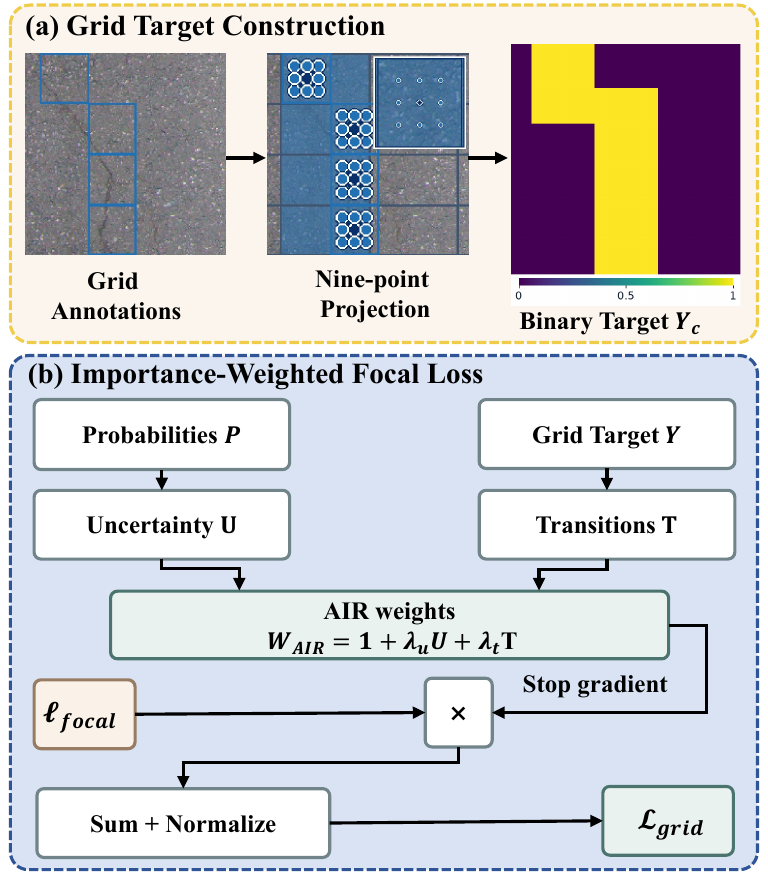}
    \caption{Grid target construction and adaptive importance reweighting.
    (a) Nine-point projection converts transformed grid annotations into
    binary category-wise targets.
    (b) Prediction uncertainty and target transitions determine detached
    weights for focal supervision over all grid entries.}
    \label{fig:supp_air_pipeline}
\end{figure}

Grid regions are stored separately from detection boxes and undergo
the same geometric transformations as the image.
Each transformed region has a grid category $c$ and normalized
coordinates $(c_x,c_y,w,h)$.
Starting from $\mathbf Y=\mathbf0$, project nine points:
\begin{equation}
    \begin{aligned}
        v&=\left\lfloor(c_x+d_x\delta_gw)W_g\right\rfloor,\\
        u&=\left\lfloor(c_y+d_y\delta_gh)H_g\right\rfloor,
    \end{aligned}
    \quad d_x,d_y\in\{-1,0,1\},
    \label{eq:box_to_grid_projection}
\end{equation}
where $\delta_g=0.125$.
Valid indices set $Y_c(u,v)=1$, and repeated projections are merged.
The implementation casts projected coordinates to integers, which
agrees with the floor expression for nonnegative projected positions.
Out-of-range indices are ignored.
Categories remain independent, and fully annotated images without
positive regions retain all-negative targets.
Fig.~\ref{fig:supp_air_pipeline} illustrates this construction and
the subsequent weighting procedure.

\subsection{Loss Formulation and Gradient Properties}
\label{supp:loss}

For probability $p$ and binary target $y$, binary focal loss is
\begin{equation}
    \begin{aligned}
        p_t&=yp+(1-y)(1-p),\\
        \alpha_t&=y\alpha_f+(1-y)(1-\alpha_f),\\
        \ell_{\mathrm{focal}}(p,y)
        &=-\alpha_t(1-p_t)^{\gamma_f}\log p_t.
    \end{aligned}
    \label{eq:binary_focal}
\end{equation}

We use $\alpha_f=0.5$, $\gamma_f=1.5$,
$\lambda_u=\lambda_t=1$, and $k_t=3$.
Logits are clipped to $[-20,20]$ before training probabilities
and focal weights are computed.
Numerically, the logarithmic term is evaluated using binary
cross-entropy with logits, multiplied by the focal and class factors.

Dilation and erosion operate independently on each category using
$k_t\times k_t$ neighborhoods, unit stride, and valid locations
at image borders.
Thus, $U_j\in[0,1]$ and $T_j\in\{0,1\}$.
The implementation also clips importance weights to $[1,10]$,
which is inactive under the stated coefficients because $1\leq W_j\leq3$.
Let $N=\max(1,N_+)$ and define
\begin{equation}
    \mathcal L_0=\frac{1}{N}\sum_j
    \ell_{\mathrm{focal}}(P_j,Y_j).
\end{equation}

Nonnegative focal losses yield
\begin{equation}
    \mathcal L_0\leq\mathcal L_{\mathrm{grid}}\leq3\mathcal L_0.
    \label{eq:air_loss_bound}
\end{equation}

Because the weights are detached and $N$ depends only on targets,
\begin{equation}
    \begin{aligned}
        \frac{\partial\mathcal L_{\mathrm{grid}}}{\partial\bar Z_j}
        &=\frac{W_j}{N}
        \frac{\partial\ell_{\mathrm{focal}}(P_j,Y_j)}{\partial\bar Z_j}\\
        &=W_j\frac{\partial\mathcal L_0}{\partial\bar Z_j}.
    \end{aligned}
\end{equation}

The resulting sign preservation applies to individual clipped logits.
For raw logits, the chain rule additionally includes the clipping
derivative, which is zero outside the clipping interval.
The property does not imply that the full parameter gradient retains
its direction or that optimization is guaranteed to converge.
Every category-cell entry remains in the loss, and setting
$\lambda_u=\lambda_t=0$ recovers $\mathcal L_0$.

\section{Detailed Inference Procedure}
\label{supp:inference}

\subsection{Decoding and Category Alignment}
\label{supp:decoding}

Let $\mathbf I_0\in\mathbb R^{3\times H_0\times W_0}$ be the
original image, and let $\mathcal T$ denote input preprocessing.
The detector's original postprocessor decodes its configured query
outputs and restores boxes to original-image coordinates.
Category maps $\kappa_d$ and $\kappa_g$ map detection labels and grid
channels to the same semantic category space.
The decoded set $\mathcal D_0$ is retained for box average precision.
Association uses the subset with scores strictly greater than $\tau_d$.

Grid probabilities are computed as $\mathbf P=\sigma(\mathbf Z)$.
Each cell selects the highest-scoring category over all output channels,
with ties resolved by the smallest channel index.
The grid threshold uses $s^g\geq\tau_g$ on RoadDistress2026,
Crack500, and DeepCrack.
Under direct resizing, cell $(u,v)$ maps to
\begin{equation}
    \mathbf b^g_{uv}
    =\left[\frac{vW_0}{W_g},\frac{uH_0}{H_g},
    \frac{(v+1)W_0}{W_g},\frac{(u+1)H_0}{H_g}\right].
    \label{eq:grid_to_image}
\end{equation}

Indices are zero-based. For other supported preprocessing, the corresponding inverse geometric mapping replaces Eq.~\ref{eq:grid_to_image}.

\subsection{Association and Complete Algorithm}
\label{supp:association}

For $\mathbf b=[x_1,y_1,x_2,y_2]$, two boxes have positive
intersection area if both intersection side lengths are positive.
Boundary touching alone does not qualify. This is equivalent to the original rule $\operatorname{IoA}(g,d)>0$, where $\operatorname{IoA}(g,d)=
|\mathbf b^g\cap\mathbf b^d|/(|\mathbf b^g|+\epsilon)$.
No grid-cell expansion or positive overlap-ratio threshold is used.

Algorithm~\ref{alg:joint_inference} specifies decoding and association
for direct resizing.
Grid-only mode returns unassociated predictions from the grid branch.
Association never substitutes a lower-scoring category when the
highest-scoring one has no matching detection.
It preserves detection outputs and all retained grid scores.
If there are no accepted detections, joint mode returns an empty
grid set, whereas grid-only mode remains independent of detections.
Box average precision is computed from $\mathcal D_0$ before
the association confidence cutoff.

\begin{algorithm}[t]
    \caption{Class-Aware Box--Grid Association}
    \label{alg:joint_inference}
    \small
    \algrenewcommand{\algorithmicrequire}{\textbf{Input:}}
    \algrenewcommand{\algorithmicensure}{\textbf{Output:}}
    \begin{algorithmic}[1]
        \Require Original image $\mathbf I_0$;
        preprocessing $\mathcal T$; model $\mathcal M$;
        thresholds $\tau_d,\tau_g$;
        category maps $\kappa_d,\kappa_g$;
        inference mode $m$.
        \Ensure Detections $\mathcal D$ and grid output
        $\mathcal G_{\mathrm{out}}$.

        \State $\mathbf I\gets\mathcal T(\mathbf I_0)$.
        \State $(\mathcal D_{\mathrm{raw}},\mathbf Z)
            \gets\mathcal M(\mathbf I)$.
        \State $\mathcal D_0\gets
            \operatorname{Decode}
            (\mathcal D_{\mathrm{raw}},\mathcal T^{-1},\kappa_d)$.
        \State $\mathcal D\gets
            \{d\in\mathcal D_0\mid s^d>\tau_d\}$.
        \State Set $\mathbf P\gets\sigma(\mathbf Z)$
            and $\mathcal G\gets\varnothing$.

        \For{$u=0,\ldots,H_g-1$}
            \For{$v=0,\ldots,W_g-1$}
                \State $c\gets\min\arg\max_j P_j(u,v)$.
                \State $s\gets P_c(u,v)$.
                \If{$s\geq\tau_g$}
                    \State Compute $\mathbf b^g$
                        using Eq.~\eqref{eq:grid_to_image}.
                    \State $g\gets
                        (\mathbf b^g,s,\kappa_g(c))$.
                    \State $\mathcal G\gets
                        \mathcal G\cup\{g\}$.
                \EndIf
            \EndFor
        \EndFor

        \If{$m=\text{grid-only}$}
            \State \Return $\mathcal D,\mathcal G$.
        \EndIf

        \State Initialize $\mathcal G^*\gets\varnothing$.
        \For{each grid candidate $g\in\mathcal G$}
            \For{each detection $d\in\mathcal D$
                with $\hat c^d=\hat c^g$}
                \State $\Delta_x\gets
                    \min(x_2^g,x_2^d)-\max(x_1^g,x_1^d)$.
                \State $\Delta_y\gets
                    \min(y_2^g,y_2^d)-\max(y_1^g,y_1^d)$.
                \If{$\Delta_x>0$ and $\Delta_y>0$}
                    \State $\mathcal G^*\gets
                        \mathcal G^*\cup\{g\}$.
                    \State \textbf{break}.
                \EndIf
            \EndFor
        \EndFor

        \State \Return $\mathcal D,\mathcal G^*$.
    \end{algorithmic}
\end{algorithm}

\section{Experiments}
\label{supp:additional_experiments}

We test whether FDAM improves grid localization for tinier crack
instances on RoadDistress2026 and thinner crack components on
Crack500 and DeepCrack.
This \emph{thin} describes narrow width in image coordinates, while
\emph{tiny} describes limited instance extent.
Stratified recall identifies missed positive grid cells within each
group. Overall precision and F1 check whether higher recall comes
at the cost of additional false positives.
A qualitative comparison further illustrates the complementary
box and grid representations.

\subsection{Annotation-Defined Stratification}
\label{supp:stratification}

RoadDistress2026 uses annotated boxes because verified masks are
unavailable. For a box of size $w\times h$ in an image of size
$W\times H$, we define its relative extent as
\begin{equation}
    e=\max\!\left(\frac{w}{W},\frac{h}{H}\right).
    \label{eq:supp_relative_extent}
\end{equation}

This measures instance extent rather than crack width.
For Crack500 and DeepCrack, we extract 8-connected components with
at least three pixels from the original ground-truth masks.
For component $C$, its relative width proxy is
\begin{equation}
    r(C)=\frac{
        \operatorname{median}_{p\in\operatorname{Skel}(C)}
        2\operatorname{EDT}_{C}(p)
    }{\min(H,W)},
    \label{eq:supp_relative_width}
\end{equation}
where $\operatorname{Skel}$ denotes skeletonization and
$\operatorname{EDT}_{C}$ is the Euclidean distance to background.
Masks are neither resized nor dilated before measurement.
This digital proxy has discretization bias: a one-pixel-wide strip
has a proxy width of two pixels. It does not measure physical width.

For each dataset, we estimate two boundaries from training annotations
alone, using percentiles at $33\frac{1}{3}\%$ and $66\frac{2}{3}\%$.
These boundaries, $q_1$ and $q_2$, remain fixed for test evaluation.
Values $x\leq q_1$, $q_1<x\leq q_2$, and $x>q_2$ define the
lower, middle, and upper groups, respectively.
We call them Tiny, Medium, and Large for instance extent, and
Thin, Medium, and Thick for the width proxy.

Equal measurements remain in the same group, so group sizes may differ.
Specifically, $r(C)\leq q_1^r$ defines the thin group on each public
dataset, while $e\leq q_1^e$ defines the tiny group on
RoadDistress2026 used to examine performance on tiny crack instances.
The superscripts indicate the measurement, and the boundaries are
estimated independently for each dataset.
These relative groups operationalize width and extent for this analysis.
They imply neither universal size thresholds nor a correspondence to
physical crack width. RoadDistress2026 boxes support extent analysis
only and cannot establish whether a crack is physically thin.

\subsection{Evaluation Protocol}
\label{supp:strata_protocol}

We retain the existing project splits, checkpoints, and fixed
inference settings. The test sets contain 1,300 RoadDistress2026,
1,124 Crack500, and 109 DeepCrack images.
RoadDistress2026 uses direct resizing to $1024\times1024$ and
a $32\times32$ grid.
FDAM uses a grid threshold of $\tau_g=0.5$ on RoadDistress2026
and $\tau_g=0.45$ on the public datasets.
GCBD uses $\tau_g=0.35$ for unassociated grids and $\tau_g=0.15$
for associated output. Both detection methods use $\tau_d=0.30$.
Grid scores satisfy $s_g\geq\tau_g$, whereas detection scores
satisfy $s_d>\tau_d$.
Association requires positive overlap with a same-category box.
On Crack500, the pixel thresholds are $0.25$ for SCSegamba
and $0.35$ for MixerCSeg. Both use $0.05$ on DeepCrack.
Segmentation probabilities are bilinearly resized to $640\times640$
and thresholded. At least $1\%$ foreground occupancy marks a cell
positive in the $20\times20$ grid.
These comparisons therefore include conversion and operating-point
differences. We report the recomputed outputs at these fixed settings.

Grid-group membership is assigned to source annotation regions
before the original nine-point target projection.
Mixed or unresolved cells, including those involving excluded
small components, are omitted only from stratified recall.
They remain in overall metrics.
The assigned proportions of positive category-cell entries are
$99.70\%$, $98.04\%$, and $98.19\%$, respectively.
Within each group, recall is the fraction of ground-truth positive
category-cell entries correctly predicted as positive.
RoadDistress2026 averages recall over longitudinal and transverse
crack categories. Public-dataset recall concerns the single crack class.
Overall RoadDistress2026 scores also average these two categories.
Overall precision and F1 include background false positives without
averaging the background class.
Box recall uses confidence-ordered, same-category matching at
$\operatorname{IoU}\geq0.50$ before grouping ground-truth instances.
It measures recall at a fixed threshold, not AP.

\begin{table*}[t]
    \centering
    \caption{Grid recall within each annotation-defined group and
    overall grid performance, in percent. RoadDistress2026 groups
    reflect instance extent, whereas public-dataset groups reflect
    relative mask width. GCBD (grid) uses unassociated predictions.
    GCBD (box--grid) and all FDAM results use class-aware box--grid
    association. The grid thresholds for these output modes are
    fixed separately as specified in Sec.~\ref{supp:strata_protocol}.
    Best and second-best metric values are \bestval{bold}
    and \secondval{underlined}, respectively.
    Rankings are computed separately within each dataset.}
    \label{tab:supp_strata_results}
    \small
    \setlength{\tabcolsep}{3pt}
    \renewcommand{\arraystretch}{1.12}
    \begin{tabularx}{\textwidth}{@{}l*{6}{C}@{}}
        \toprule
        \rowcolor{FDAMHeader}
        & \multicolumn{3}{c}{Recall within each group}
        & \multicolumn{3}{c}{Overall grid performance} \\
        \cmidrule(lr){2-4}\cmidrule(lr){5-7}
        \multicolumn{7}{l}{\textbf{RoadDistress2026}} \\
        \rowcolor{FDAMHeader}
        Method & Tiny & Medium & Large & Precision & Recall & F1 \\
        \midrule
        GCBD (grid) & \secondval{39.32} & \secondval{58.07} & \secondval{61.47} & \secondval{48.77} & \secondval{58.61} & \secondval{52.95} \\
        GCBD (box--grid) & 6.53 & 23.24 & 31.93 & 44.33 & 26.67 & 32.96 \\
        \textbf{FDAM (Ours)} & \bestval{44.50} & \bestval{60.90} & \bestval{72.25} & \bestval{61.28} & \bestval{67.51} & \bestval{64.24} \\
        \midrule
        \multicolumn{7}{l}{\textbf{Crack500}} \\
        \rowcolor{FDAMHeader}
        Method & Thin & Medium & Thick & Precision & Recall & F1 \\
        \midrule
        SCSegamba & \secondval{73.72} & 84.44 & \secondval{86.03} & \secondval{83.51} & \secondval{83.84} & \secondval{83.67} \\
        MixerCSeg & 72.71 & \secondval{84.79} & 85.69 & 82.31 & 83.59 & 82.95 \\
        \textbf{FDAM (Ours)} & \bestval{85.17} & \bestval{88.16} & \bestval{88.28} & \bestval{84.89} & \bestval{87.84} & \bestval{86.34} \\
        \midrule
        \multicolumn{7}{l}{\textbf{DeepCrack}} \\
        \rowcolor{FDAMHeader}
        Method & Thin & Medium & Thick & Precision & Recall & F1 \\
        \midrule
        SCSegamba & \secondval{70.84} & \secondval{81.01} & 83.29 & \bestval{91.83} & 80.50 & \secondval{85.79} \\
        MixerCSeg & 67.67 & 80.53 & \secondval{84.91} & 84.61 & \secondval{80.64} & 82.58 \\
        \textbf{FDAM (Ours)} & \bestval{86.45} & \bestval{88.43} & \bestval{90.64} & \secondval{89.35} & \bestval{89.43} & \bestval{89.39} \\
        \bottomrule
    \end{tabularx}
\end{table*}

\subsection{Stratified Results}
\label{supp:strata_results}

\begin{figure}[!t]
    \centering
    \includegraphics[width=\linewidth]{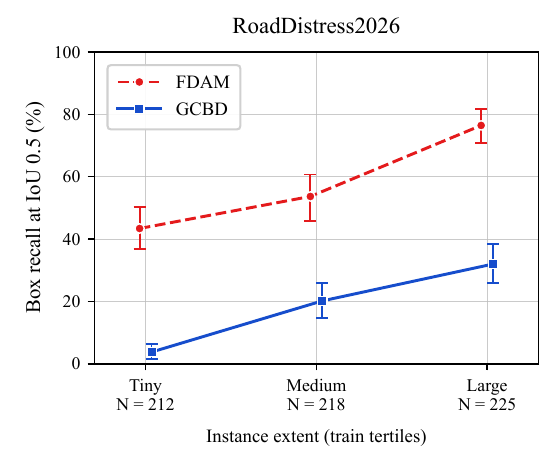}
    \caption{RoadDistress2026 box recall across relative instance
    extents at score $>0.30$ and $\operatorname{IoU}\geq0.50$.
    $N$ counts annotated instances within each group.
    Error bars denote 95\% image-bootstrap intervals.}
    \label{fig:supp_box_strata}
\end{figure}

On RoadDistress2026, associated FDAM exceeds GCBD's separately
selected grid output by $5.17$ percentage points in small-extent
grid recall.
Fig.~\ref{fig:supp_box_strata} additionally shows small-extent
box recall of $43.40\%$ versus $3.77\%$.
These results concern instance extent, not actual crack width.
However, association reduces FDAM's overall grid F1 from
$64.53\%$ to $64.24\%$ and small-extent grid recall from
$46.97\%$ to $44.50\%$.

\begin{figure*}[!tp]
    \centering
    \setlength{\abovecaptionskip}{4pt}
    \setlength{\belowcaptionskip}{0pt}
    \includegraphics[width=\textwidth]{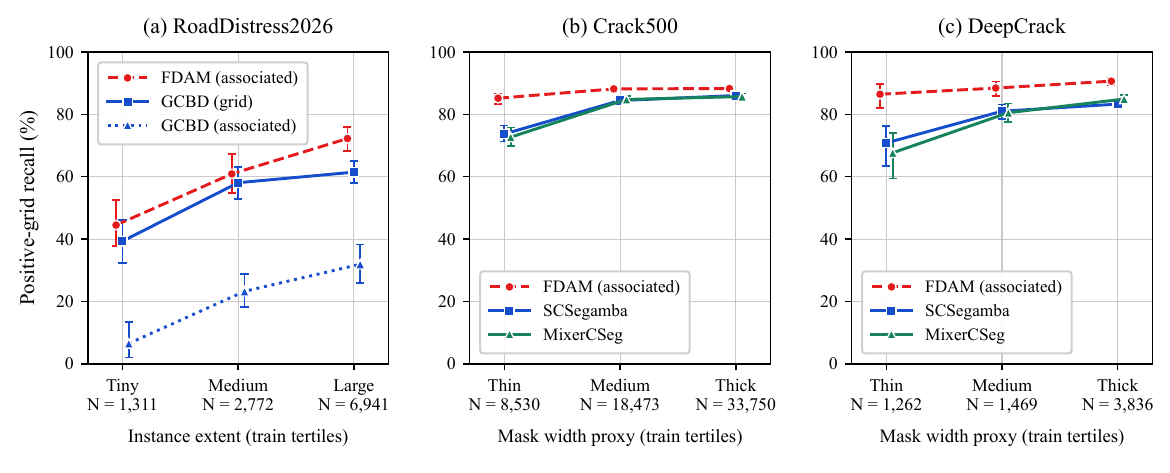}
    \caption{Grid recall across annotation-defined strata.
    RoadDistress2026 uses relative instance extent, whereas Crack500
    and DeepCrack use the ground-truth mask width proxy.
    The protocols are specified in Sec.~\ref{supp:strata_protocol}.
    $N$ counts positive category-cell entries within each group.
    Error bars denote 95\% image-bootstrap intervals.}
    \label{fig:supp_grid_strata}

    \vspace{8pt}

    \includegraphics[width=\textwidth]{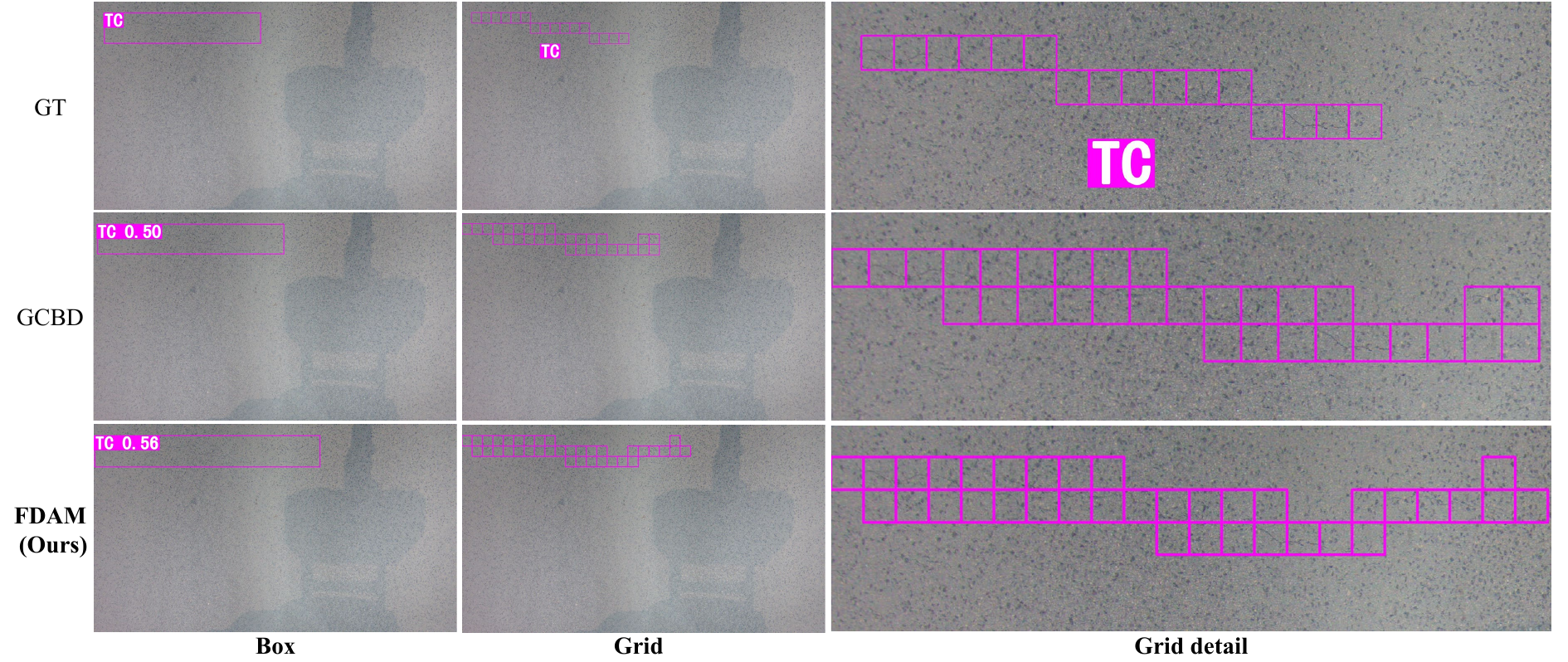}
    \caption{Qualitative comparison of pavement
    crack grid localization on RoadDistress2026.
    Rows show GT, GCBD predictions, and FDAM predictions.
    Columns show box labels or predictions, grid labels
    or predictions, and enlarged views of the grid results.
    Magenta denotes transverse cracks (TC).
    The enlarged views facilitate inspection of local
    crack occupancy.}
    \label{fig:supp_road_comparison}
\end{figure*}

FDAM preserves more positive grid responses in the relative Thin
groups of both public datasets.
As shown in Fig.~\ref{fig:supp_grid_strata}, its gains over SCSegamba
are $11.45$ percentage points on Crack500 and $15.61$ on DeepCrack.
These point-estimate gains exceed those in the medium and thick groups.
Overall foreground F1 is also higher, although DeepCrack precision
is lower than SCSegamba's.
These observations do not establish statistical significance or
attribute the gains to individual modules.

\subsection{Qualitative Comparison}
\label{supp:road_qualitative}

Fig.~\ref{fig:supp_road_comparison} compares GCBD and FDAM
on a transverse-crack example from RoadDistress2026.
Box predictions describe the overall instance extent,
while grid predictions represent local crack occupancy.
The enlarged grid views expose differences in cell coverage
that are difficult to inspect in the full images.
This example illustrates the complementary spatial
representations used for pavement crack grid localization.
Stratified quantitative results are reported in
Table~\ref{tab:supp_strata_results}, and broader dataset-wide
comparisons are provided in the main paper.

\end{document}